\documentclass{article}

\usepackage[final]{corl_2025}

\usepackage{bm}
\usepackage{svg}
\usepackage{xcolor}
\usepackage{amssymb}
\usepackage{caption}
\usepackage{amsmath}
\usepackage{wrapfig}
\usepackage{graphicx}
\usepackage{geometry}
\usepackage{multirow}
\usepackage{booktabs}
\usepackage{algorithm}
\usepackage{subcaption}
\usepackage{algpseudocode}

\usepackage{enumitem}

\graphicspath{{figures/}}

\newcommand{\CostFunc}{g}
\newcommand{\Node}{\mathcal{N}}
\newcommand{\ObsHistory}{\bm{o}} 
\newcommand{\PlanIndices}{\bm{b}}
\newcommand{\DeltaQ}{\Delta\bm{q}}
\newcommand{\PlanSet}{\mathcal{P}}
\newcommand{\FrontierSet}{\mathcal{F}}
\newcommand{\CollisionCache}{\mathcal{C}}
\newcommand{\PairedObsHistory}{\hat{\bm{o}}}
\newcommand{\DualAgentModel}{\bm{\epsilon}_{\theta_2}}
\newcommand{\SingleAgentModel}{\bm{\epsilon}_{\theta_1}}

\newcommand{\DualAgentModelQL}{\bm{\epsilon}_{\theta_2}^{\texttt{QL}}}

\newcommand{\SingleAgentModelQL}{\bm{\epsilon}_{\theta_1}^{\texttt{QL}}}

\title{Diffusion-Guided Multi-Arm Motion Planning}

\author{
  Viraj Parimi\thanks{Correspondence to \texttt{vparimi@mit.edu}.}\;, Brian Williams \\
  Massachusetts Institute of Technology \\
}

\begin{document}
\maketitle


\begin{abstract}
    Multi-arm motion planning is fundamental for enabling arms to complete complex long-horizon tasks in shared spaces efficiently but current methods struggle with scalability due to exponential state-space growth and reliance on large training datasets for learned models. Inspired by Multi-Agent Path Finding (MAPF), which decomposes planning into single-agent problems coupled with collision resolution, we propose a novel diffusion-guided multi-arm planner (DG-MAP) that enhances scalability of learning-based models while reducing their reliance on massive multi-arm datasets. Recognizing that collisions are primarily pairwise, we train two conditional diffusion models, one to generate feasible single-arm trajectories, and a second, to model the dual-arm dynamics required for effective pairwise collision resolution. By integrating these specialized generative models within a MAPF-inspired structured decomposition, our planner efficiently scales to larger number of arms. Evaluations against alternative learning-based methods across various team sizes demonstrate our method's effectiveness and practical applicability. Project website: \url{https://diff-mapf-mers.csail.mit.edu}
\end{abstract}



\section{Introduction}
\label{sec:introduction}

The ability of multiple arms to effectively coordinate in shared spaces is crucial for solving complex tasks beyond the capability of individual arms. Humans naturally perform many such sophisticated tasks by leveraging social interaction \cite{shared-intentionality}, shared information \cite{SEBANZ200670}, and distributed responsibilities \cite{teams}. Enabling robotic systems, particularly those with multiple manipulators operating in close proximity, to achieve similar efficient, collision-free coordination remains a central challenge. This involves handling high-dimensional joint configuration spaces inherent to multi-arm systems. Traditional sampling-based motion planners (SMPs), such as RRT and PRM \cite{lavalle2006planning, LaValle1998RapidlyexploringRT, rrtstar, rrtconnect, Qureshi_2015, prm}, face significant challenges when applied to multi-arm systems due to the curse of dimensionality in the joint space. Alternatively, optimization-based planners can efficiently refine trajectories but are susceptible to local minima and often require good initial seeds. Recently, learning-based methods have emerged that can guide such planners \cite{seo2025prestofastmotionplanning, huang2024diffusionseederseedingmotionoptimization} with a primary focus on enhancing single-arm planning. Extending them to multi-arm setups would require expensive multi-arm training data and solving large, non-convex optimization problems. To address the multi-agent scalability more directly, other methods  \cite{chen2022cooperativetaskmotionplanning} have integrated SMPs with Multi-Agent Path Finding (MAPF) decomposition techniques \cite{mapf_benchmarks}, constructing individual roadmaps for each arm and coordinating them using MAPF-like solvers. Nonetheless, scalability remains constrained, as the computational cost of generating initial SMP roadmaps becomes a major bottleneck, particularly as the number of arms increases.

To address these scalability challenges, end-to-end learning-based approaches have gained increasing attention. In particular, Multi-Agent Reinforcement Learning (MARL) methods \cite{ha2020multiarm} have emerged, training decentralized policies using expert demonstrations from SMPs such as BiRRT. While they have been shown to be scalable, MARL policies often depend heavily on diverse training data, and those trained primarily on simpler interactions involving only two arms frequently fail to generalize to more complex, larger team scenarios. 

An alternative line of work leverages diffusion models \cite{song2021scorebasedgenerativemodelingstochastic, ho2020denoisingdiffusionprobabilisticmodels, carvalho2024motionplanningdiffusionlearning, janner2022planningdiffusionflexiblebehavior, zhu2025madiffofflinemultiagentlearning}, which offer a promising combination of generative flexibility and explicit constraint handling. For instance, approaches such as Multi-robot Multi-model planning Diffusion (MMD) \cite{shaoul2024multirobotmotionplanningdiffusion} integrate single-robot diffusion models within MAPF frameworks to enable collision-free navigation. Nonetheless, encoding complex multi-agent geometric constraints, particularly for articulated manipulators, within diffusion models remains a substantial challenge. Furthermore, current end-to-end multi-agent diffusion policies \cite{zhu2025madiffofflinemultiagentlearning} require centralized training with full-team data, limiting their scalability to unseen team compositions and motivating the need for novel, multi-arm-specific approaches.

\begin{figure*}[t]
    \centering  
    \includegraphics[width=\textwidth]{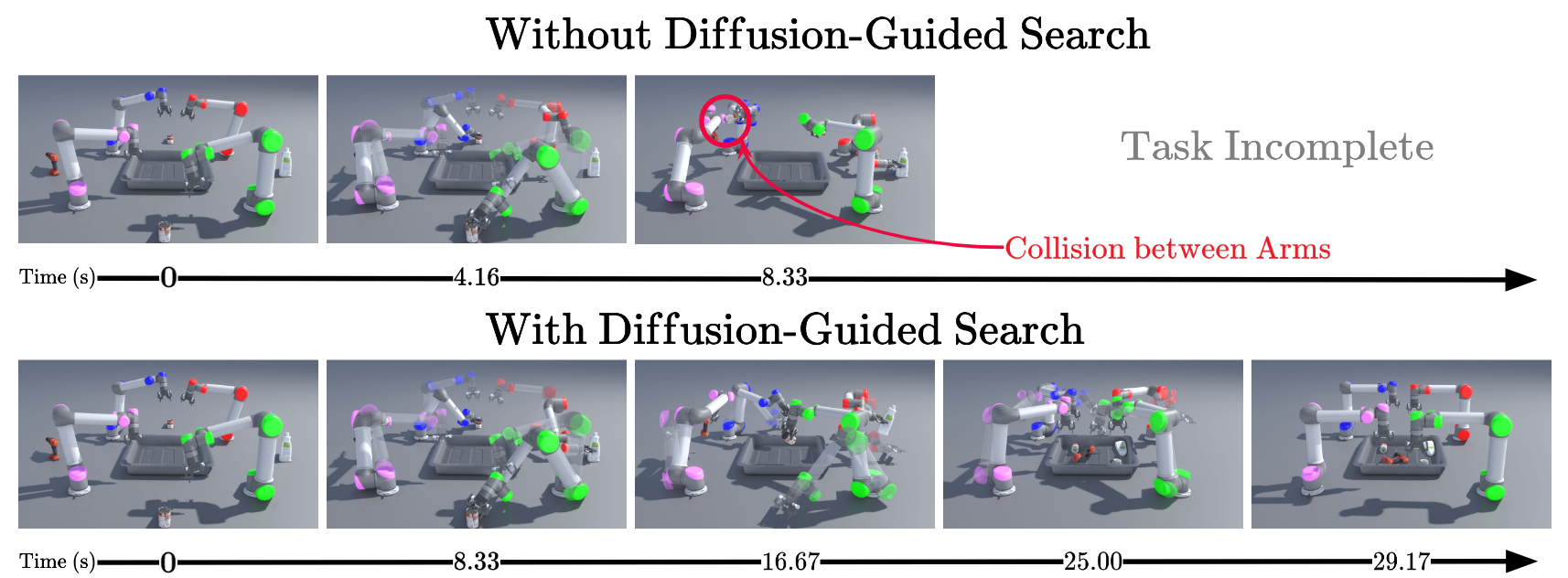}
    \caption{\textbf{Multi-Arm Motion Planning}: Timestamped snapshots compare an end-to-end learning method (top) with our DG-MAP approach (bottom) on a multi-arm pick-and-place task. Both were trained using only lower-order interaction data (single/dual-arm trajectories). The end-to-end method fails due to collision with another arm in the shared workspace, while DG-MAP successfully completes the task by leveraging specialized diffusion models combined with MAPF-inspired structured decomposition.}
    \label{fig:compare}
\end{figure*}

Achieving effective multi-arm coordination requires planners that are scalable to large teams, adaptable to dynamic layouts, cooperative in goal achievement, and capable of closed-loop operation for collision avoidance, as highlighted by Ha et al. \cite{ha2020multiarm} and illustrated in Figure \ref{fig:compare}. Motivated by these challenges, we propose a novel closed-loop multi-arm motion planning framework called DG-MAP that combines structured decomposition principles from MAPF with conditional diffusion models to achieve scalable and data-efficient planning. Instead of planning directly in the high-dimensional joint space, we plan in the \emph{conflict-resolution space} wherein we plan motions for each arm independently and use search to iteratively deconflict these plans from any potential collisions among pairs of arms using conflict-resolution based methods. To this end, we introduce two specialized conditional denoising diffusion models. The first model learns to generate feasible single-arm trajectories respecting individual constraints. The second model is specifically designed to generate feasible trajectories necessary to effectively resolve pairwise conflicts, conditioned on the relevant states of the interacting pair. By integrating these generative models within a MAPF-inspired framework, our planner manages combinatorial complexity and reduces the dependency on large-scale higher-order multi-arm training data. 

Our main contributions are:
\begin{itemize}[itemsep=0.1pt]
    \item Conditional denoising diffusion models tailored for generating single-arm trajectories and, distinctively, for resolving pairwise arm collisions.
    \item A novel framework integrating specialized diffusion models with MAPF-inspired decomposition for multi-arm motion planning.
    \item Empirical validation showcasing significant improvements in scalability and effectiveness even when trained on simpler interaction scenarios compared to alternative approaches.
\end{itemize}

\section{Preliminaries}
\label{sec:probdef}

\subsection{Problem Definition}

Consider a total of $N$ manipulator arms, where each arm $i$ has a fixed base pose $\bm{\xi}_i^{\text{base}}$, $d_i$ degrees-of-freedom (DoF), configuration space $\mathcal{Q}_i \subseteq \mathbb{R}^{d_i}$, and collision-free subspace $\mathcal{Q}_i^{\text{free}}$ with respect to itself and any static obstacles in the environment. Multi-arm motion planning seeks a simultaneous continuous path $\bm{\tau}: [0, T] \to \mathcal{Q}_{\text{sys}}$, where $\mathcal{Q}_{\text{sys}} = \times_{i=1}^N \mathcal{Q}_i$ and $T > 0$ is the final time when all arms have reached their goals, mapping time $t$ to the system configuration $\bm{q}(t) = (\bm{q}_1(t), \dots, \bm{q}_N(t)) = \bm{\tau}(t) \in \mathcal{Q}_i$. The path $\bm{\tau}$ must satisfy:
\begin{enumerate}[leftmargin=*,align=left]
    \item \textbf{Boundary Conditions:} $\bm{\tau}(0) = \bm{q}_{\text{start}}$ and $\bm{\tau}(T) = \bm{q}_{\text{final}}$ such that for all $i$, the final end-effector pose $\bm{p}_i^{\text{ee}}(\bm{q}_i(T)) = \texttt{FK}(\bm{q}_i(T))$ is within tolerances ($\delta_{\text{pos}}, \delta_{\text{rot}}$) of the goal $\bm{p}_i^{\text{goal}}$, i.e., $d_{\text{pos}}(\bm{p}_i^{\text{ee}}(\bm{q}_i(T)), \bm{p}_i^{\text{goal}}) \le \delta_{\text{pos}}$ and $d_{\text{rot}}(\bm{p}_i^{\text{ee}}(\bm{q}_i(T)), \bm{p}_i^{\text{goal}}) \le \delta_{\text{rot}}$.
    \item \textbf{Collision Avoidance:} For all $t \in [0, T]$, each arm is collision-free ($\bm{q}_i(t) \in \mathcal{Q}_i^{\text{free}} \; \forall \; i$) and there are no inter-arm collisions between any distinct pair $(i, j)$, $i \neq j$.
\end{enumerate}

\subsection{Multi-Agent Path Finding (MAPF)}

Multi-Agent Path Finding (MAPF) \cite{mapf_benchmarks} is a discrete abstraction of the multi-arm coordination problem that seeks collision-free paths $\bm{\tau}$ for $N$ agents on a graph representing feasible configurations and transitions. Scalable constraint-based algorithms like Prioritized Planning (PP) \cite{prioriplan} and Conflict-Based Search (CBS) \cite{sharon-cbs} excel in MAPF by decomposing the multi-agent planning problem into a series of single-agent problems. They rely on single-agent planners to propose individual trajectories for each agent independent of each other and resolve detected conflicts between two agents by imposing explicit spatio-temporal constraints on subsequent planning queries.

\subsection{Diffusion Models}

Diffusion models, particularly Denoising Diffusion Probabilistic Models (DDPMs) \cite{ho2020denoisingdiffusionprobabilisticmodels, song2021scorebasedgenerativemodelingstochastic}, are probabilistic generative models that learn to reverse a fixed Markovian process that gradually adds Gaussian noise to data over $K$ steps. To generate a sample $\bm{z}_0$, such as a trajectory segment, the model starts from pure noise $\bm{z}_K \sim \mathcal{N}(0, \mathbf{I})$ and iteratively denoises it via:
\begin{equation}
\label{eq:ddpm_reverse}
    \bm{z}_{k-1} = f(\bm{z}_k, \bm{\epsilon}_{\theta}(\bm{z}_k, k)) \quad \text{for } k = K, \dots, 1,
\end{equation}
where $\bm{\epsilon}_{\theta}$ is a neural network predicting the noise at each step, and $f(\cdot)$ denotes the denoising update based on the noise schedule. Training minimizes the mean squared error between the true and predicted noise. For a data point $\bm{z}_0$ and a timestep $k \sim \text{Uniform}(1, K)$, noise $\bm{\epsilon} \sim \mathcal{N}(0, \mathbf{I})$ is added according to, $\bm{z}_k = \sqrt{\bar{\alpha}_k}\bm{z}_0 + \sqrt{1 - \bar{\alpha}_k}\bm{\epsilon}$, where $\bar{\alpha}_k = \prod_{i=1}^k \alpha_i$ is the cumulative noise schedule. The loss function then minimizes the difference between the predicted and true noise:
\begin{equation}
\label{eq:ddpm_loss}
    \mathcal{L}(\theta) = \mathbb{E}_{\bm{z}_0, \bm{\epsilon}, k} \left[ \| \bm{\epsilon} - \bm{\epsilon}_{\theta}(\bm{z}_k, k) \|^2 \right].
\end{equation}

Due to their ability to model complex, multi-modal distributions and their stable training dynamics, diffusion models have been successfully applied to robotics tasks such as trajectory planning \cite{janner2022planningdiffusionflexiblebehavior, carvalho2024motionplanningdiffusionlearning, chi2023diffusionpolicy} making them suitable for generating meaningful manipulator trajectories. 

\section{Approach}
\label{sec:approach}

We propose DG-MAP, our scalable and closed-loop multi-arm motion planner that integrates specialized generative models like conditional denoising diffusion models within a MAPF-inspired structured decomposition. First, we discuss details about the offline training of these specialized diffusion models following which we will present a simple search-based strategy to integrate these models into a closed-loop planning framework.

\begin{figure*}[t]
    \centering  
    \includegraphics[width=\textwidth]{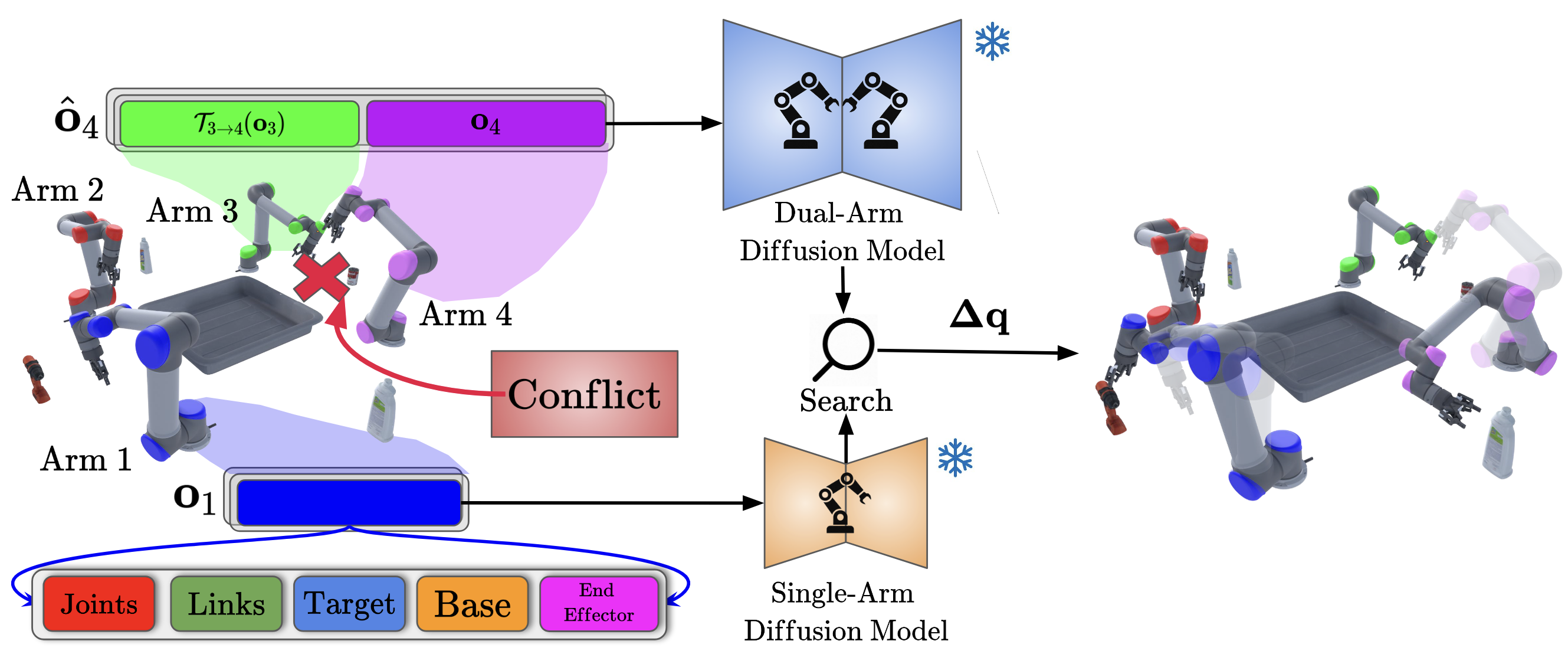}
    \caption{Overview of how DG-MAP operates in a closed-loop, receding horizon controller, generating delta-action sequences for multiple arms. Initially, each arm plans independently using its dedicated single-arm model ($\SingleAgentModel$). Conflicts trigger Rebranch (Alg. ~\ref{alg:sub_rebranch}) Repair (Alg. ~\ref{alg:sub_repair}) strategies. The Repair strategy is highlighted (top) where a conflicting arm (pink) extracts paired observations with the other conflicting arm (green) and utilizes the trained dual-arm model ($\DualAgentModel$) to generate feasible delta-actions avoiding the collision.}
    \label{fig:overview}
\end{figure*}

\subsection{Single-Arm Diffusion Model}

Addressing the need for a closed-loop control that is crucial for multi-arm coordination with moving obstacles like other arms, we leverage the Diffusion Policy framework \cite{chi2023diffusionpolicy} to train a generative model for single-arm trajectory generation. A shared set of model parameters, denoted $\theta_1$, is learned using expert demonstrations collected from single-arm BiRRT plans, following the setup in \cite{ha2020multiarm}. During planning and execution, each arm $i$ utilizes an independent instantiation of this model to predict a trajectory tailored to its current observation history.

The policy, denoted $\bm{\epsilon}_{\theta_1}$, generates a sequence of future delta joint actions $\Delta\bm{q}_i = [\Delta\bm{q}_i(t), \dots, \Delta\bm{q}_i(t+T_p-1)]$ over a prediction horizon $T_p$. Each predicted trajectory is conditioned on a recent sequence of $T_o$ observations $\bm{o}_i = [\bm{o}_i(t-T_o-1), \dots, \bm{o}_i(t)]$ unique to arm $i$. Each observation frame $\bm{o}_i(t)$ includes the joint configuration $\bm{q}_i(t)$, end-effector pose $\bm{p}_i^{\text{ee}}(t)$, target end-effector pose $\bm{p}_i^{\text{goal}}$, link positions $\bm{L}_i(t)$, and constant base pose $\bm{\xi}_i^{\text{base}}$. This stacked observation sequence provides temporal and goal-directed context to the policy.

\begin{equation}
\label{eq:single_agent_delta_loss_explicit_obs}
\mathcal{L}(\theta_1) = \mathbb{E}_{k, \Delta\bm{q}_i^0, \bm{o}_i,  \bm{\epsilon}} \left[ \left\| \bm{\epsilon} - \bm{\epsilon}_{\theta_1}(\bm{o}_i, \Delta\bm{q}_i^k, k) \right\|^2 \right]
\end{equation}

The model $\bm{\epsilon}_{\theta_1}$ is trained via the standard diffusion objective $\mathcal{L}(\theta_1)$ (Eq.~\eqref{eq:single_agent_delta_loss_explicit_obs}) to predict the noise $\bm{\epsilon}$ added to ground-truth actions $\Delta\bm{q}_i^0$, conditioned on observations $\bm{o}_i$ and timestep $k$. After training, each arm independently samples its own action trajectory by running the denoising process using its current observation history $\bm{o}_i$. This trained model serves to generate initial, potentially conflicting, trajectory proposals for individual arms. 

\subsection{Dual-Arm Diffusion Model}

To resolve inter-arm conflicts during multi-arm planning, we introduce a dual-arm diffusion model $\bm{\epsilon}_{\theta_2}$ that learns using expert demonstrations collected from dual-arm BiRRT plans. While the action space remains unchanged where we are predicting delta joint action sequences $\Delta\bm{q}_i$ for the ego-arm, the observation structure is reorganized to explicitly capture interactions between the ego-arm and the conflicting arm from this interacting pair inspired from \cite{SEBANZ200670, ha2020multiarm}.

The dual-arm observation, denoted as $\hat{\bm{o}}_i$, is constructed by \textit{pairing} the transformed observations of the conflicting arm with the ego-arm's own observations at each timestep across a history window of $T_o$ steps. Specifically, for each timestep $t' \in [{t-T_o-1, \dotsc, t}]$, we define $\hat{\bm{o}}_i(t') = \left[ \mathcal{T}_{j \to i}(\bm{o}_j(t')) \oplus \bm{o}_i(t') \right]$, where $\mathcal{T}{j \to i}(\cdot)$ denotes the transformation that maps arm $j$'s observations into arm $i$'s reference frame, and $\oplus$ represents concatenation. The full observation input to the diffusion model is then $\hat{\bm{o}}_i = \left[ \hat{\bm{o}}_i(t-T_o-1), \dotsc, \hat{\bm{o}}_i(t) \right]$. The dual-arm diffusion model $\bm{\epsilon}_{\theta_2}$ is trained to predict the added noise at each diffusion timestep $k$, minimizing the denoising loss:
\begin{equation}
\label{eq:dual_agent_loss}
    \mathcal{L}(\theta_2) = \mathbb{E}_{k, \Delta\bm{q}_i^0, \hat{\bm{o}}_i, \bm{\epsilon}} \left[ \left\| \bm{\epsilon} - \bm{\epsilon}_{\theta_2}(\hat{\bm{o}}_i, \Delta\bm{q}_i^k, k) \right\|^2 \right],
\end{equation}
where $\Delta\bm{q}_i^0$ denotes the ground truth sequence of delta joint actions. The model thus learns to generate collision-avoiding ego-arm trajectories while conditioning on the relevant information of the conflicting arm.

\subsection{Diffusion-Guided Multi-Arm Planning (DG-MAP)}

We integrate the trained single-arm ($\SingleAgentModel$) and dual-arm ($\DualAgentModel$) diffusion models within a search framework inspired by MAPF principles. The planner searches over candidate delta-action sequences for each arm, leveraging the diffusion models for proposal generation and conflict resolution. The overall procedure is detailed in Algorithm~\ref{alg:dg_search}, utilizing the subroutines defined in Algorithms~\ref{alg:sub_rebranch} and \ref{alg:sub_repair}.

\setlength\intextsep{0pt}
\begin{wrapfigure}[34]{r}{0.6\textwidth}
\begin{minipage}{0.58\textwidth}
\begin{algorithm}[H]
\caption{Diffusion-Guided Multi-Arm Planner (DG-MAP)}
\label{alg:dg_search}
    \begin{algorithmic}[1]
    \State \textbf{Input:} Models $\SingleAgentModel, \DualAgentModel$
    \State \textbf{Output:} Collision-free plan $\{\DeltaQ_i\}_{i=1}^N$ or best effort
    \State \textbf{Initialize:} Frontier set $\FrontierSet \gets \emptyset$, collision cache $\CollisionCache \gets \emptyset$
    \For{$i = 1 \dotsc N$}
        \State $\ObsHistory_i \gets \texttt{GetObs(i)}$
        \State $\PlanSet_i \gets \SingleAgentModel(\ObsHistory_i, \cdot, \cdot)$
        \label{line:main_init_plans}
    \EndFor
    \Statex \hspace{\algorithmicindent} \Comment{Generate initial plans independent of each} 
    \Statex \hspace{\algorithmicindent} \quad \quad \;\; \color{blue}\text{other}\color{black}
    \State $\Node_0 \gets \texttt{Node}(\bm{b} = \vec{0}, \mathcal{K} = \emptyset)$
    \Statex \hspace{\algorithmicindent} \Comment{Node stores indices and conflicting plan}
    \Statex \hspace{\algorithmicindent} \quad \quad \quad \;\; \color{blue}\text{indices}\color{black}
    \State $\FrontierSet.\texttt{Insert}(\Node_0, \CostFunc(\Node_0))$
    
    \While{$\FrontierSet$ not empty and time not exceeded} \label{line:main_while_start_explicit}
        \State $\Node \leftarrow \FrontierSet.\texttt{ExtractMin}()$
        \State $\bm{\tau} \gets \{ \bm{\Delta q}_i^{\Node.b_i} \}_{i=1}^N$
        \State $\bm{c} \gets \texttt{FindFirstCollision}(\bm{\tau}, \mathcal{C})$ \label{line:main_find_collision}
        \If{$\bm{c}$ is \textbf{null}} \label{line:main_solution_check}
            \State \Return $\bm{\tau}, t^*$ \Comment{Solution found}
        \EndIf
        \State $(i, j, \hat{t}) \gets \bm{c}$ \Comment{Conflicting pair and time}
        \State $t^* = \min(t^*, \hat{t})$ \Comment{Update earliest collision time}
        \State $\kappa_i = \Node.\mathcal{K} \; \cup \Node.b_i$ \Comment{Attempt to fix for arm $i$}
        \State $\texttt{Rebranch}(i, \kappa_i)$ \label{line:call_rebranch_i}
        \State $\texttt{Repair}(i, j, \kappa_i)$ \label{line:call_repair_i}
        \State $\kappa_j = \Node.\mathcal{K} \; \cup \Node.b_j$ \Comment{Attempt to fix for arm $j$}
        \State $\texttt{Rebranch}(j, \kappa_j)$ \label{line:call_rebranch_j}
        \State $\texttt{Repair}(j, i, \kappa_j)$ \label{line:call_repair_j}
        
    \EndWhile \label{line:main_while_end_explicit}
    \State \Return Best plan found in $\FrontierSet$ based on cost
    \Statex \hspace{\algorithmicindent} \Comment{Timeout or failure}
    \end{algorithmic}
\end{algorithm}
\end{minipage}
\end{wrapfigure}
\setlength\intextsep{12pt}
\noindent\textbf{Planner State and Initialization:} The search state is represented by nodes $\Node$ in a search tree, each corresponding to a tuple of selected plan indices $\PlanIndices = (b_1, \dots, b_N)$. An index $b_i \in \{1, \dots, B\}$ points to a candidate delta-action sequence $\DeltaQ_i^{b_i}$ for arm $i$, covering a prediction horizon $T_p$. The search maintains a frontier set $\FrontierSet$, implemented as a min-priority queue storing nodes $\Node$ ordered by a cost function $\CostFunc(\Node)$ guiding the search towards promising nodes. This cost estimates solution quality by combining path smoothness, goal proximity and penalty for collisions. Initially, each arm $i$ uses its single-arm model $\SingleAgentModel$ conditioned on its observation history $\ObsHistory_i$ of up to $T_o$ steps  in the past to sample $B$ diverse candidate delta-action sequences $\PlanSet_i = \{\DeltaQ_i^1, \dots, \DeltaQ_i^B\}$. The search begins with a root node $\Node_0$ placed in $\FrontierSet$ along with an empty set of conflicting plan indices ($\mathcal{K}$) that will be updated upon expansions. A collision cache $\CollisionCache$ stores pairwise collision check results.

\begin{figure}[!t]
\begin{minipage}{0.48\textwidth}
\begin{algorithm}[H]
\caption{Generate Successors by Rebranch}
\label{alg:sub_rebranch}
    \begin{algorithmic}[1]
    \State \textbf{Input:} $\texttt{ego} \text{ arm}, \text{conflicts } \kappa$
    \For{$m = 1 \dotsc |\PlanSet_{\texttt{ego}}|$}
        \If{$m \notin \kappa$}
            \State $b_{\texttt{ego}}^{\text{new}} \gets m$
            \State $\PlanIndices' \gets (b_1, \dotsc, b_{\texttt{ego}}^{\text{new}}, \dotsc, b_N)$
            \State $\Node' \gets \texttt{Node}(\bm{b} = \PlanIndices', \mathcal{K} = \kappa)$
            \State $\FrontierSet.\texttt{Insert}(\Node', \CostFunc(\Node'))$
        \EndIf
    \EndFor
    \end{algorithmic}
\end{algorithm}
\end{minipage}
\hfill
\begin{minipage}{0.48\textwidth}
\begin{algorithm}[H]
\caption{Generate Successors by Repair}
\label{alg:sub_repair}
    \begin{algorithmic}[1]
    \State \textbf{Input:} \parbox[t]{\dimexpr\linewidth-\algorithmicindent}{%
        $\texttt{ego}$ arm, $\texttt{other}$ arm, conflicts $\kappa$,\\
        Model $\DualAgentModel$
    }
    \State $\PairedObsHistory_{\texttt{ego}} \gets \texttt{GetPairedObs}(\texttt{ego}, \texttt{other})$
    \State Sample $\{\DeltaQ_{\texttt{ego}}^{\text{new},m}\}_{m=1}^B$ using $\DualAgentModel(\PairedObsHistory_{\texttt{ego}}, \cdot, \cdot)$
    \For{$m = 1 \dotsc B$}
        \State $b_{\texttt{ego}}^{\text{new}} \gets \PlanSet_{\texttt{ego}}.\texttt{Update}(\DeltaQ_{\texttt{ego}}^{\text{new}, m})$
        \State $\PlanIndices' \gets (b_1, \dotsc, b_{\texttt{ego}}^{\text{new}}, \dotsc, b_N)$
        \State $\Node' \gets (\PlanIndices = \PlanIndices', \mathcal{K} = \kappa)$
        \State $\FrontierSet.\texttt{Insert}(\Node', \CostFunc(\Node'))$
    \EndFor
    \end{algorithmic}
\end{algorithm}
\end{minipage}
\end{figure}

\noindent\textbf{Search Process and Conflict Resolution:} The main loop (lines~\ref{line:main_while_start_explicit}-\ref{line:main_while_end_explicit}) iteratively extracts the lowest-cost node $\Node$ from $\FrontierSet$. It checks the corresponding plan combination for collisions using \texttt{FindFirstCollision} (line~\ref{line:main_find_collision}). If no conflict is found, a valid solution is returned (line~\ref{line:main_solution_check}).

As the node $\Node$ is expanded we detect a conflict $c = (i, j, \hat{t})$, where $\hat{t}$ is the earliest time step within the prediction horizon $T_p$ where the plans $\DeltaQ_i^{b_i}$ and $\DeltaQ_j^{b_j}$ collide. $t^*$ is the earliest conflict time and is updated based on $\hat{t}$. It signifies that the current combination of plans is collision-free up to time $t^*-1$, which could potentially allow for safe partial execution in a receding horizon context, although our primary goal here is to find a fully collision-free plan for the horizon $T_p$. The set of conflicts $\kappa_i$ indicate the plan indices corresponding to arm $i$ that have been found to be in conflict in the ancestors of the node. This book-keeping helps in avoiding redundant plans that did not succeed in earlier expansions. The node expansion generates successor nodes by exploring alternatives for the conflicting arms $i$ and $j$ using two distinct strategies (lines~\ref{line:call_rebranch_i}-\ref{line:call_repair_j}). First, \texttt{Rebranch} (Algorithm~\ref{alg:sub_rebranch}) creates successors by trying alternative, pre-existing plans from $\PlanSet$ for both $i$ and $j$. Second, \texttt{Repair} (Algorithm~\ref{alg:sub_repair}) uses the dual-arm model $\DualAgentModel$ to generate $B$ new candidate sequences for both $i$ and $j$, specifically aiming to avoid the detected conflict $c$, and creates successors for each successful repair. All generated successors are added to $\FrontierSet$, allowing the search to explore other alternatives based on their estimated costs.

\noindent\textbf{Termination:} The search continues until a collision-free node is extracted or termination conditions (timeout or empty frontier) are met, returning the best plan found.

DG-MAP (Algorithm~\ref{alg:dg_search}) operates within a closed-loop, receding horizon controller (Figure \ref{fig:overview}). At each step, it plans from the current state, returning action sequences $\{\DeltaQ_i^{b_i}\}_{i=1}^N$ for horizon $T_p$ and the predicted collision-free duration $t^* \le T_p$. Actions are executed for $t^*$ steps, the state is updated, and the process repeats. This interleaved planning and execution ensures adaptation, persistent collision avoidance, and progress towards goals $\{\bm{p}_i^{\text{goal}}\}_{i=1}^N$. Executing up to $t^*$ provides a balance between reactivity and planning efficiency, avoiding the excessive planning time of single-step execution.

\section{Experiments}
\label{sec:evaluation}

Our experiments aim to answer the following key questions regarding DG-MAP:
\begin{itemize}[noitemsep, leftmargin=*, align=left]
    \item[\textbf{Q1:}] How effectively does DG-MAP scale to planning for larger numbers of arms ($N=3$ to $N=8$) when trained only on single and dual-arm interaction data?
    \item[\textbf{Q2:}] How does DG-MAP's performance compare to the learning-based methods, trained on limited interaction data?
    \item[\textbf{Q3:}] Can DG-MAP remain competitive with the learning-based methods, which leverage significantly more complex higher-order interaction data during training?
    \item[\textbf{Q4:}] Can DG-MAP be applied to practical applications apart from simple single goal-reaching tasks? 
\end{itemize}

\begin{figure*}[ht]
    \centering
    \begin{subfigure}{0.32\textwidth}
        \includegraphics[width=\linewidth]{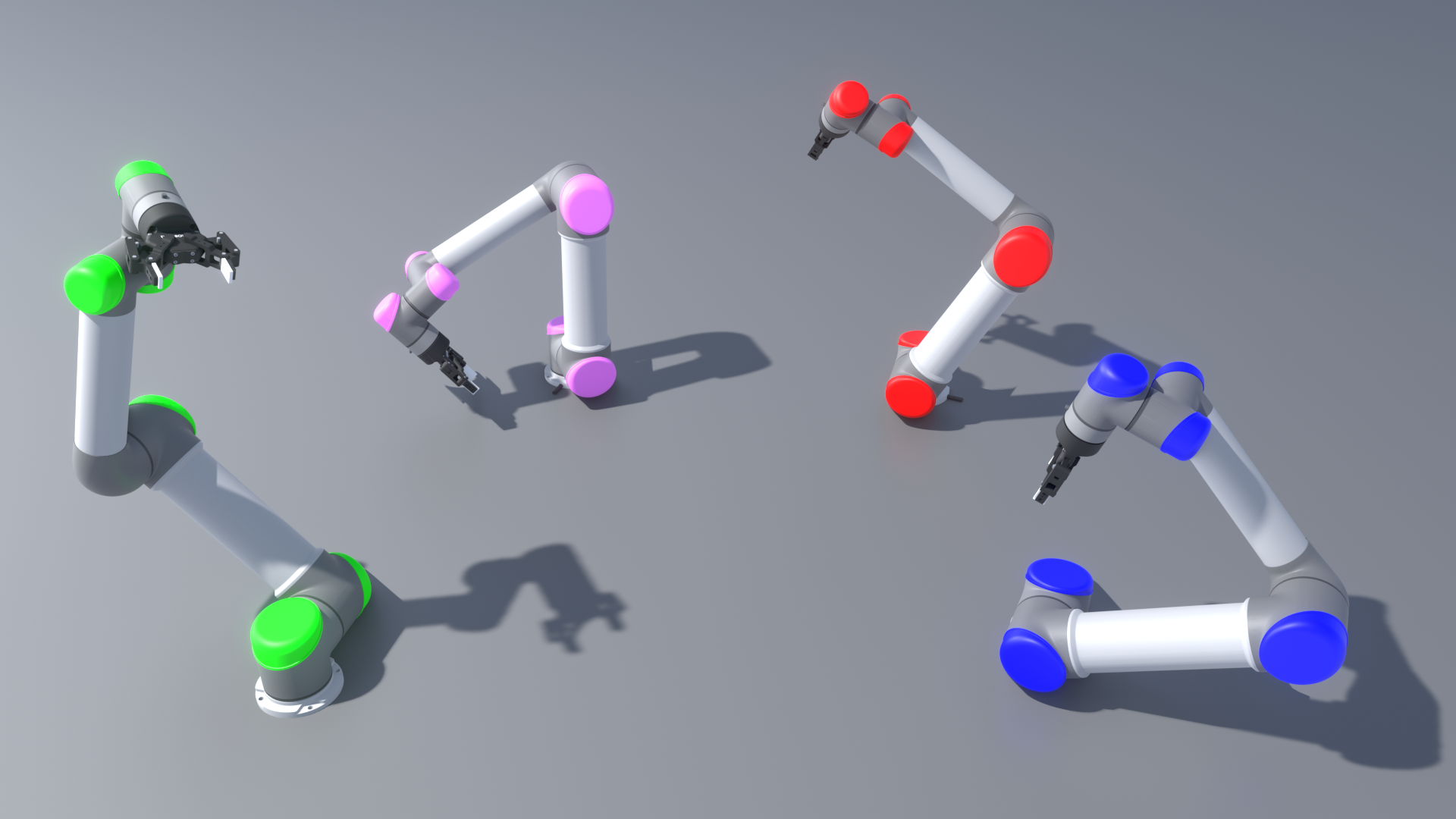}
    \end{subfigure}
    \hfill
    \begin{subfigure}{0.32\textwidth}
        \includegraphics[width=\linewidth]{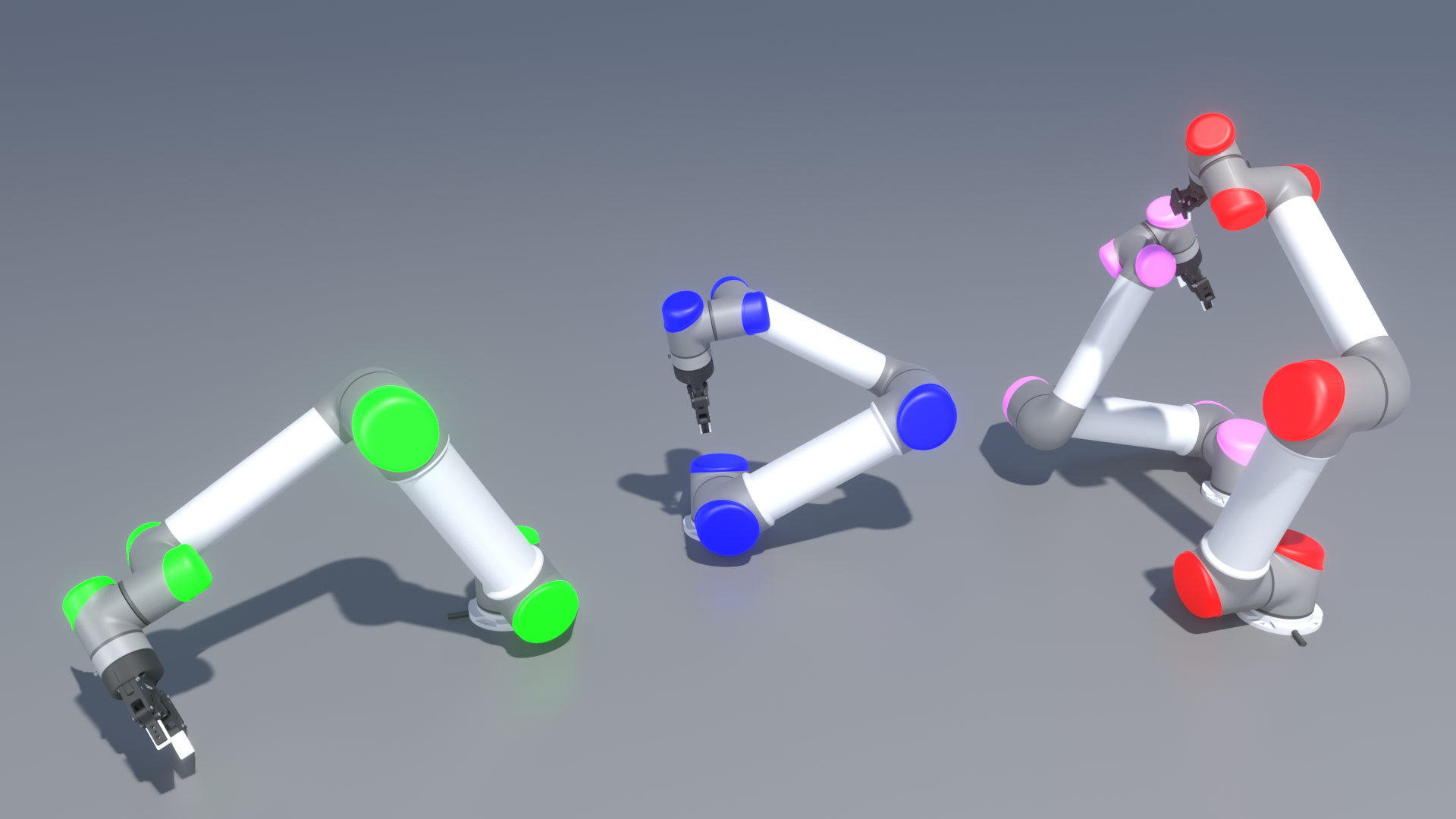}
    \end{subfigure}
    \hfill
    \begin{subfigure}{0.32\textwidth}
        \includegraphics[width=\linewidth]{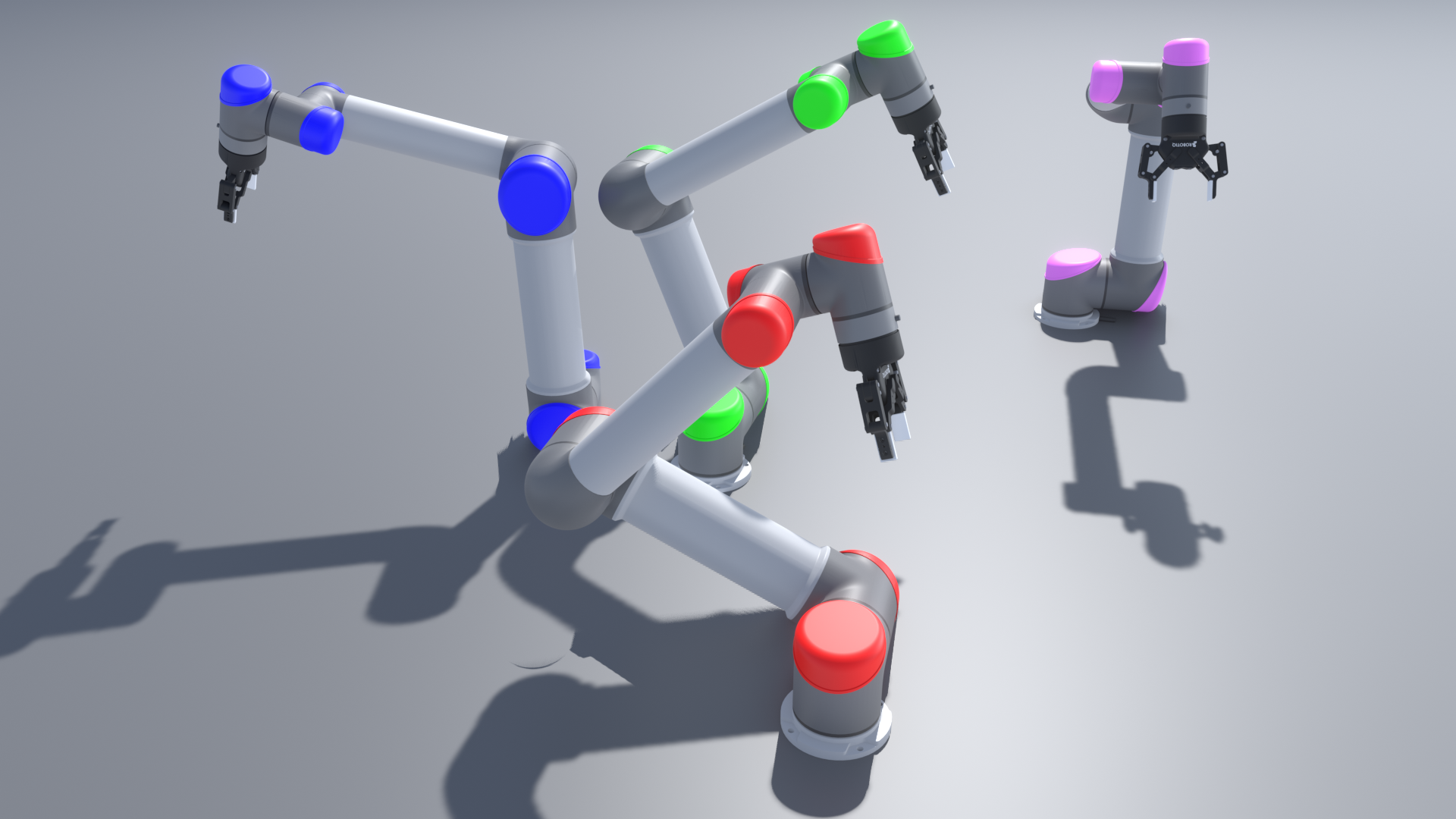}
    \end{subfigure}
    \caption{Visualization of test tasks for four arms across different difficulty levels: easy (left), medium (middle), and hard (right). Task difficulty is based on the maximum intersection between the arms' hemispherical workspaces \cite{ha2020multiarm}, with easy tasks involving minimal arm interaction and hard tasks requiring significant overlap and coordination.}
    \label{fig:task_types}
\end{figure*}

\noindent\textbf{Setup:} We focus on static single goal-reaching tasks with $N = 3$ to $N = 8$ manipulator arms, forming a foundation for applications like pick-and-place. Each arm has $d_i = 6$ DoF. Tasks are executed in the PyBullet simulator \cite{coumans2016pybullet} and are considered successful if all $N$ arms reach their target end-effector poses $\bm{p}_i^{\text{goal}}$ within a positional tolerance $\delta_{\text{pos}} = 0.03$ units and orientation tolerance $\delta_{\text{rot}} = 0.1$ radians. Each attempt is limited to 400 simulation steps. We generate a test dataset of 18,000 unique scenarios, with novel base poses $\bm{\xi}_i^{\text{base}}$ and target poses $\bm{p}_i^{\text{goal}}$ distinct from training data, evenly distributed across arm counts. Following \cite{ha2020multiarm}, tasks are categorized by difficulty based on the maximum workspace intersection volume between pairs of arms, as illustrated in the Figure \ref{fig:task_types}. Implementation specific details can be found in the Appendix.

For comparisons, we utilize two variants of the multi-arm motion planner from Ha et al. \cite{ha2020multiarm}, which represents a valuable learning-based baseline with demonstrable scalability assurances.
\begin{enumerate}[noitemsep, leftmargin=*, align=left]
    \item \textbf{Baseline-LD (Limited Data):} This variant is trained using only single-arm and dual-arm interactions for up to 756.8M timesteps. This comparison directly assesses the effectiveness of our approach given the same limited interaction data.
    \item \textbf{Baseline-ED (Extended Data):} This variant is trained on extended interaction including three and four arms for up to 668.2 M timesteps allowing us to quantify the performance differential between models that can leverage complex multi-arm coordination compared to simplistic pairwise interactions combined with planning.
\end{enumerate}
\vspace{-0.5em}

\noindent\textbf{Comparison with Baseline-LD:} Table~\ref{tab:baselineld} shows that DG-MAP significantly outperforms Baseline-LD across all task difficulties and team sizes. While Baseline-LD achieves moderate success with 3 arms (41.2\% on average), its performance quickly deteriorates with more arms, dropping below 10\% and sometimes failing entirely on medium and hard tasks beyond 4 arms. In contrast, DG-MAP maintains consistently high average success rates above 90\% across all settings. This stark contrast highlights the scalability of DG-MAP's planning-guided coordination in complex multi-arm environments, where learning-only methods like Baseline-LD struggle due to limited training on higher-order interactions addressing both $\textbf{Q1}$ and $\textbf{Q2}$.
\begin{table}[ht]
\resizebox{\textwidth}{!}{%
\begin{tabular}{@{}ccccccccc@{}}
\toprule
\multirow{2}{*}{\textbf{Arms}} &
  \multicolumn{2}{c}{\textbf{Easy}} &
  \multicolumn{2}{c}{\textbf{Medium}} &
  \multicolumn{2}{c}{\textbf{Hard}} &
  \multicolumn{2}{c}{\textbf{Average}} \\ \cmidrule(l){2-9} 
 &
  \begin{tabular}[c]{@{}c@{}}Baseline\\ LD\end{tabular} &
  \begin{tabular}[c]{@{}c@{}}DG-MAP\\ Ours ($\uparrow$) \end{tabular} &
  \begin{tabular}[c]{@{}c@{}}Baseline\\ LD\end{tabular} &
  \begin{tabular}[c]{@{}c@{}}DG-MAP\\ Ours ($\uparrow$) \end{tabular} &
  \begin{tabular}[c]{@{}c@{}}Baseline\\ LD\end{tabular} &
  \begin{tabular}[c]{@{}c@{}}DG-MAP\\ Ours ($\uparrow$) \end{tabular} &
  \begin{tabular}[c]{@{}c@{}}Baseline\\ LD\end{tabular} &
  \begin{tabular}[c]{@{}c@{}}DG-MAP\\ Ours ($\uparrow$) \end{tabular} \\ \midrule
3 & 0.349 & \textbf{0.984} & 0.426 & \textbf{0.975} & 0.460 & \textbf{0.965} & 0.412 & \textbf{0.975} \\
4 & 0.032 & \textbf{0.981} & 0.024 & \textbf{0.969} & 0.060 & \textbf{0.953} & 0.039 & \textbf{0.968} \\
5 & 0.224 & \textbf{0.972} & - & \textbf{0.958} & - & \textbf{0.905} & 0.075 & \textbf{0.945}  \\
6 & 0.145 & \textbf{0.976} & 0.011 & \textbf{0.921} & 0.008 & \textbf{0.888} & 0.055 & \textbf{0.928} \\
7 & 0.113 & \textbf{0.955} & - & \textbf{0.918} & - & \textbf{0.907} & 0.038 & \textbf{0.926} \\
8 & 0.067 & \textbf{0.951} & - & \textbf{0.933} & - & \textbf{0.888} & 0.022 & \textbf{0.924} \\ \bottomrule
\end{tabular}%
}
\caption{Success rates (\%, higher is better) for Baseline-LD and DG-MAP across easy, medium, and hard tasks, evaluated with 3 to 8 arms. - indicates no successes.}
\label{tab:baselineld}
\end{table}

\noindent\textbf{Comparison with Baseline-ED:} Table~\ref{tab:baselineed} presents success rates for Baseline-ED compared to DG-MAP. While Baseline-ED maintains high performance across all settings (over 85\% success), DG-MAP consistently outperforms it with substantial gains for larger teams, with relative improvements up to 3.8\% on medium-difficulty tasks with 6 arms and 3.3\% on hard tasks with 8 arms. This demonstrates DG-MAP’s planning-guided coordination provides robust scalability and efficiency even in dense, high-interaction scenarios, making it competitive even with end-to-end learned policies that have been trained on higher-order interaction data answering $\textbf{Q3}$.
\begin{table}[!t]
\resizebox{\textwidth}{!}{%
\begin{tabular}{@{}ccccccccc@{}}
\toprule
\multirow{2}{*}{\textbf{Arms}} &
  \multicolumn{2}{c}{\textbf{Easy}} &
  \multicolumn{2}{c}{\textbf{Medium}} &
  \multicolumn{2}{c}{\textbf{Hard}} &
  \multicolumn{2}{c}{\textbf{Average}} \\ \cmidrule(l){2-9} 
 &
\begin{tabular}[c]{@{}c@{}}Baseline\\ ED\end{tabular} &
  \begin{tabular}[c]{@{}c@{}}DG-MAP\\ Ours ($\uparrow$) \end{tabular} &
  \begin{tabular}[c]{@{}c@{}}Baseline\\ ED\end{tabular} &
  \begin{tabular}[c]{@{}c@{}}DG-MAP\\ Ours ($\uparrow$) \end{tabular} &
  \begin{tabular}[c]{@{}c@{}}Baseline\\ ED\end{tabular} &
  \begin{tabular}[c]{@{}c@{}}DG-MAP\\ Ours ($\uparrow$) \end{tabular} &
  \begin{tabular}[c]{@{}c@{}}Baseline\\ ED\end{tabular} &
  \begin{tabular}[c]{@{}c@{}}DG-MAP\\ Ours ($\uparrow$) \end{tabular} \\ \midrule
3 & 0.980 & \textbf{0.984} & 0.973 & \textbf{0.975} & 0.943 & \textbf{0.965} & 0.965 & \textbf{0.975} \\
4 & 0.973 & \textbf{0.981} & 0.961 & \textbf{0.969} & 0.950 & \textbf{0.953} & 0.961 & \textbf{0.968} \\
5 & 0.963 & \textbf{0.972} & 0.947 & \textbf{0.958} & 0.891 & \textbf{0.905} & 0.934 & \textbf{0.945} \\
6 & 0.950 & \textbf{0.976} & 0.887 & \textbf{0.921} & 0.882 & \textbf{0.888} & 0.906 & \textbf{0.928} \\
7 & 0.950 & \textbf{0.955} & 0.913 & \textbf{0.918} & 0.891 & \textbf{0.907} & 0.917 & \textbf{0.926} \\
8 & \textbf{0.951} & \textbf{0.951} & 0.911 & \textbf{0.933} & 0.859 & \textbf{0.888} & 0.907 & \textbf{0.924} \\ \bottomrule
\end{tabular}%
}
\caption{Success rates (\%, higher is better) for Baseline-ED and DG-MAP across easy, medium, and hard tasks, evaluated with 3 to 8 arms.}
\label{tab:baselineed}
\end{table}

\noindent\textbf{Multi-Arm Pick-And-Place Task}

\begin{figure}[ht]
\begin{minipage}{0.48\textwidth}
\centering
\includegraphics[width=\linewidth]{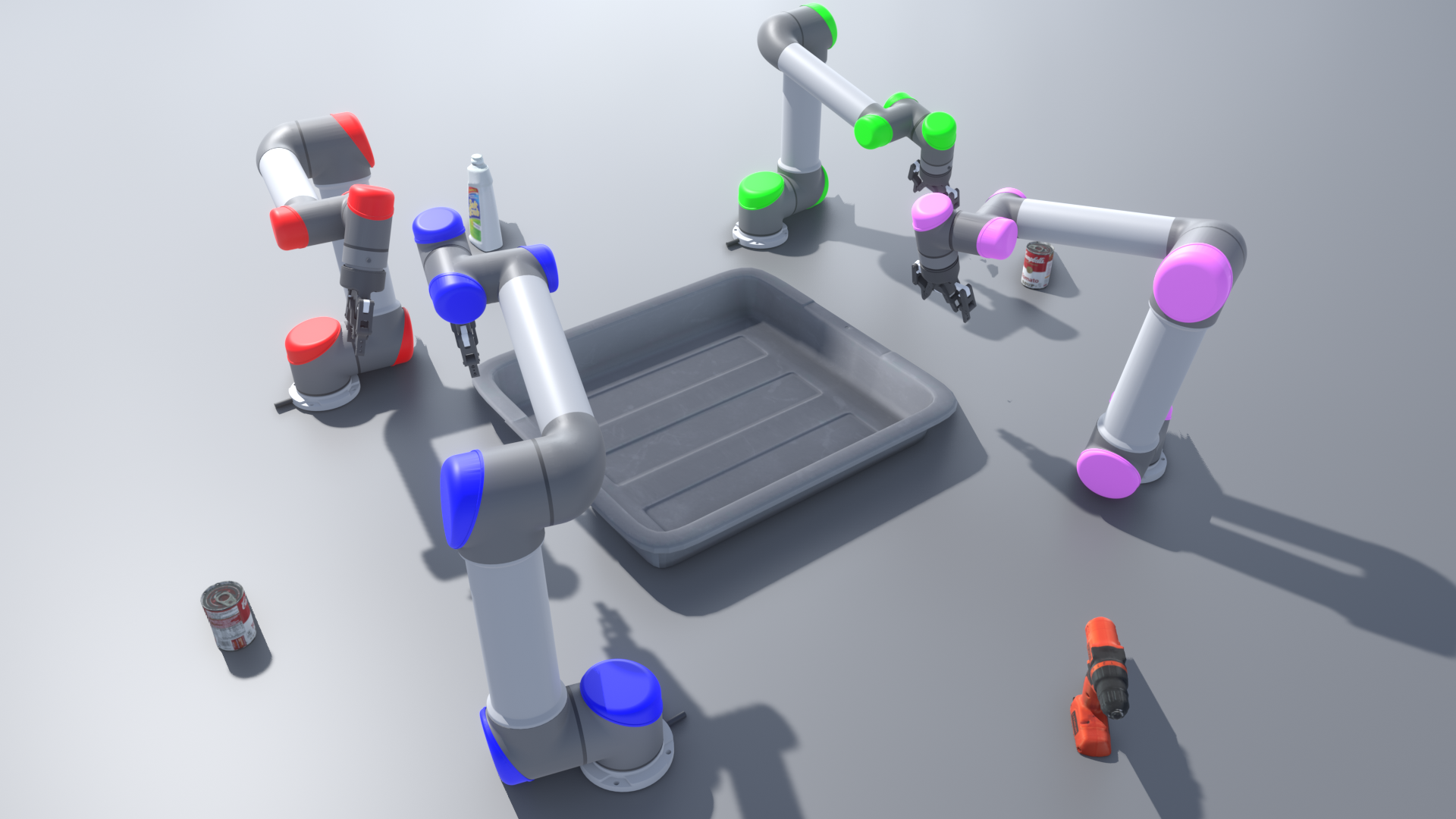}
\captionof{figure}{Visualization of the multi-arm pick-and-place task where arms are tasked to pick objects from the ground onto the central bin.}
\label{fig:picknplace}
\end{minipage}
\hfill
\begin{minipage}{0.48\textwidth}
\centering
\begin{tabular}{@{}ccc@{}}
  \toprule
  \textbf{Method} & \textbf{Success} $(\uparrow)$ & \textbf{Steps} $(\downarrow)$ \\
  \midrule
  Baseline-LD     & 0.375 & \textbf{4479} \\
  Baseline-ED     & 0.714 & 6018 \\
  DG-MAP & \textbf{0.890} & 5390 \\
  \bottomrule
\end{tabular}
\captionof{table}{Success rate (\%, higher is better) along with average number of steps (lower is better) taken by the methods to complete the full cycle of multi-arm pick-and-place task.}
\label{tab:picknplace_table}
\end{minipage}
\end{figure}

To address $\textbf{Q4}$, we evaluate DG-MAP on a four-arm, 6-DoF pick-and-place simulation setup adapted from \cite{ha2020multiarm}. In this setup, four UR5 robots equipped with Robotiq 2F-85 grippers are positioned at the corners of a central bin. The task requires the arms to collaboratively pick objects from the ground and deposit them into the bin as shown in Figure.~\ref{fig:picknplace}. Each trial involves randomly sampled grasp objects from a 7-object subset of the YCB dataset \cite{ycb}. The motion consists of two collision-free phases starting from the initial pose to a pregrasp pose and from grasp to dump pose. Grasp execution is managed by a separate low-level controller using precomputed grasps generated from GraspIt! \cite{graspit} with added noise \cite{Weisz2012PoseER}. Following \cite{ha2020multiarm}, success is defined as all arms completing the pick-and-place cycle without inter-arm or ground collisions. Bin collisions and knocked-over objects are discarded, consistent with the training dataset that only incorporates arm states without the bin or objects.

\setlength\intextsep{15pt}

Using this setup, DG-MAP is able to improve upon both Baseline-LD and Baseline-ED. Across 100 trials, DG-MAP achieves a success rate of 89\% highlighting its effectiveness in enabling scalable, collision-free multi-arm coordination even in dense, shared workspaces.

\vspace{-1em}
\section{Conclusion}
\label{sec:conclusions}
\vspace{-1em}
This paper introduced DG-MAP, a diffusion-guided multi-arm motion planner addressing the scalability challenge in multi-arm motion planning. By integrating MAPF principles with specialized single-arm ($\SingleAgentModel$) and dual-arm ($\DualAgentModel$) conditional diffusion models trained only on corresponding interaction data, DG-MAP efficiently coordinates multiple arms without requiring complex higher-order interaction data. Our experiments demonstrated that DG-MAP outperforms alternative learning-based baseline trained on identical limited data, achieving high success rates $(\textgreater 88\%)$ for up to eight arms. Furthermore, it remained competitive with a baseline trained on richer multi-arm data, highlighting the data efficiency and effectiveness of the structured pairwise resolution approach. The successful application to a complex pick-and-place task further validated its practical utility beyond simple goal-reaching tasks.

\acknowledgments{We gratefully acknowledge support from Defence Science and Technology Agency, Singapore. The authors also thank Annabel Gomez for her extensive feedback on the paper draft. The views and conclusions contained in this document are those of the authors and should not be interpreted as representing the official policies, either expressed or implied, of the sponsoring organizations or agencies.}

\bibliography{references}

\clearpage

\appendix

\section{Appendix}

\subsection{Limitations}

While DG-MAP demonstrates significant promise and effectiveness across various multi-arm tasks, several aspects offer avenues for future research and enhancement. Currently, the approach leverages forward simulation of the predicted plans to check for collisions implying that the performance can be limited and highly complex environments might challenge real-time execution. Furthermore, the training relies on low-level state information like joint values and link positions. This results in models specialized to the specific arm morphologies used during training, ensuring high precision for those systems when transferring them to real manipulators but limiting direct transferability to different robot arms or heterogeneous setups. Future work could explore incorporating morphology-agnostic representations, potentially through visual perception or with VLMs \cite{pgvlm2024} to enhance generalization across platforms.

\begin{wrapfigure}{r}{0.5\textwidth}
    \centering
    \includegraphics[width=0.48\textwidth]{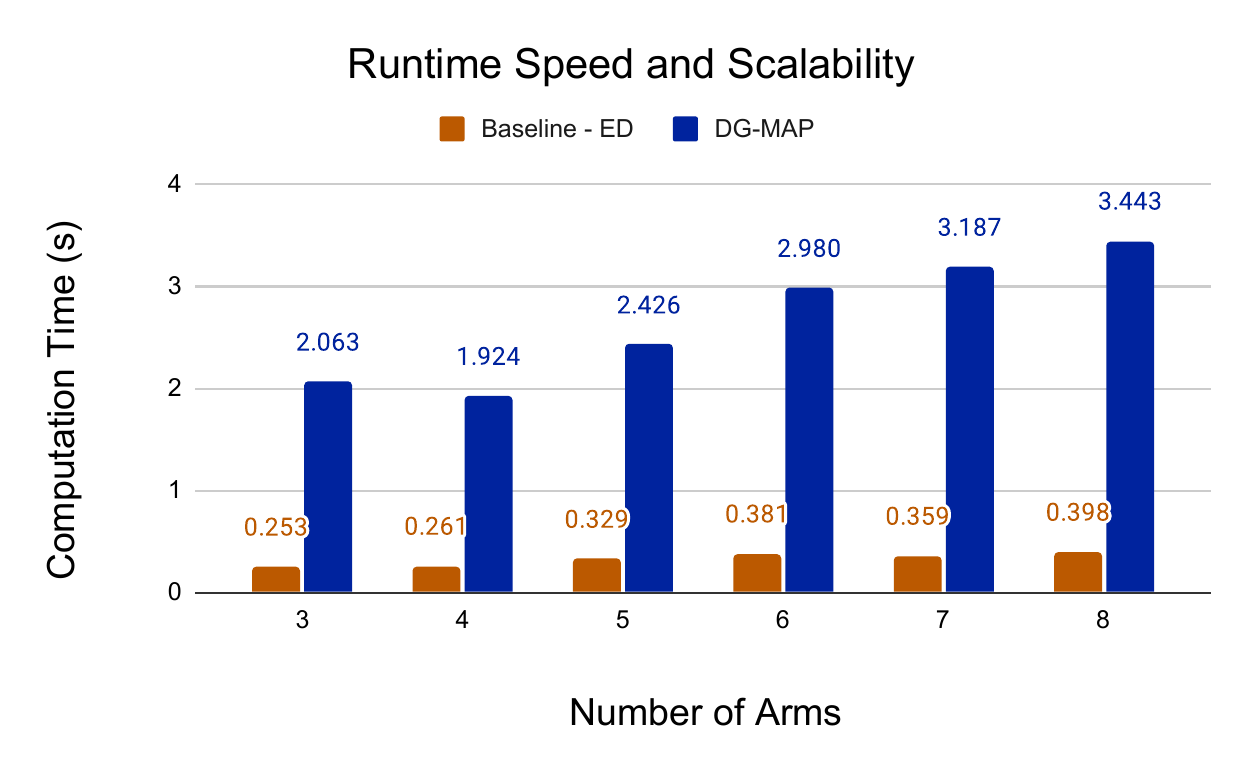}
    \caption{Computation time of Baseline-ED compared to DG-MAP in seconds}
    \label{fig:latency}
\end{wrapfigure}

The core design of DG-MAP focuses on decomposing complexity via single and dual-arm interactions, which significantly reduces training data requirements and directly addresses the most frequent conflict types. However, complex coordination strategies involving three or more arms simultaneously might not be fully captured by this pairwise approach. Investigating methods to efficiently incorporate or adaptively switch to higher-order interaction models, perhaps when pairwise resolution fails or when richer data is available, could further improve performance in highly complex scenarios.  Additionally, a key limitation is the planner's dependence on the single-arm model ($\SingleAgentModel$). If all initial trajectory proposals contain intrinsic collisions like with itself or the plane, the system cannot recover, as the dual-arm model ($\DualAgentModel$) exclusively addresses inter-arm conflicts. Finally, the closed-loop, interleaved planning and execution cycle provides reactivity but introduces computational latency during planning, as illustrated in Figure~\ref{fig:latency}. While acceptable for some robotic tasks with less outside interventions, applications demanding extremely high-speed reactions to unpredictable events, such as close human interaction, might require complementary approaches. Finally, although DG-MAP focuses its efforts on coordinating multiple arms effectively, there is still potential improvement in solving collaborative tasks as described in \cite{chen2022cooperativetaskmotionplanning}. These points highlight opportunities to build upon the DG-MAP framework, extending its applicability and performance range in future iterations.

\subsection{Implementation Details}

The diffusion models ($\SingleAgentModel$ and $\DualAgentModel$) were trained, and all planning experiments conducted, on a system equipped with a 32-core Intel i9-14900K CPU and an NVIDIA GeForce RTX 4090 GPU. The algorithm is implemented in Python and used PyBullet \cite{coumans2016pybullet} as the simulator, building upon the official codebase released by Ha et al. \cite{ha2020multiarm} for environment simulation and integrating the diffusion policy framework from Chi et al. \cite{chi2023diffusionpolicy}.

The observation vector for the single-arm model ($\SingleAgentModel$) comprised 6 joint values, 7 end-effector pose values (position + quaternion), 7 target end-effector pose values, 30 values representing key link positions, and 7 values for the fixed base pose, resulting in a dimension $|o_i| = 57$ \cite{ha2020multiarm}. The dual-arm model ($\DualAgentModel$) used a concatenated observation from the interacting pair, resulting in dimension $|\hat{o}_i| = 114$. We adopted the CNN-based UNet architecture with FiLM conditioning as detailed in \cite{chi2023diffusionpolicy}, employing the squared cosine noise schedule proposed in iDDPM \cite{nichol2021improveddenoisingdiffusionprobabilistic}.

Within the DG-MAP search (Algorithm~\ref{alg:dg_search}), nodes represent combined states for all $N$ arms. The quality of expanding a node with a specific combination of candidate plan segments $\{\DeltaQ_i^{b_i}\}_{i=1}^N$, where $\DeltaQ_i^{b_i}$ is the $b_i$-th plan segment for arm $i$ over horizon $T_p$ is evaluated using a cost function as defined below,
\[
    g(\mathcal{N}) = \sum_{i=1}^N \| \Delta\bm{q}_i^{b_i}(t) \|_2 + d_{\text{pos}}(\bm{p}_i^{\text{ee}}(\bm{q}_i(T_p-1)), \bm{p}_i^{\text{goal}}) + d_{\text{rot}}(\bm{p}_i^{\text{ee}}(\bm{q}_i(T_p - 1)), \bm{p}_i^{\text{goal}}) + P_{\text{coll}},
\]
where $\|\Delta\bm{q}_i^{b_i}(t)\|_2$ is the action magnitude, $d_{\text{pos}}(\bm{p}_i^{\text{ee}}(\bm{q}_i(T_p-1)), \bm{p}_i^{\text{goal}})$ is the position residual and $d_{\text{rot}}(\bm{p}_i^{\text{ee}}(\bm{q}_i(T_p - 1)), \bm{p}_i^{\text{goal}})$ is the orientation residual at the end of the segment $T_p$, and $P_{\text{coll}}$ is a large penalty applied if *any* collision occurs within the combined plan segment over the horizon $[0, T_p)$. Specific hyperparameter values used are listed in Table~\ref{tab:hyperparameters}.

\begin{table*}[ht]
  \centering
  \caption{Hyperparameters for DG-MAP Training and Planning.}
  \label{tab:hyperparameters}
  \begin{tabular}{@{}lr@{}}
    \toprule
    \textbf{Parameter} & \textbf{Value}                                                     \\ \midrule
    \multicolumn{2}{@{}l}{\textit{Diffusion Model \& Training}}                             \\
    Positional encoding size                                            & 256               \\
    Number of denoising steps                                           & 100               \\
    UNet layers (channels)                                              & [256, 512, 1024]  \\
    Training epochs                                                     & 100               \\
    Batch size                                                          & 4096              \\
    Observation horizon $T_o$                                           & 2                 \\
    Prediction horizon $T_p$                                            & 16                \\
    Action sequence length $T_a$ (used for training)                    & 1                 \\
    $\SingleAgentModel$ Observation dim $|o_i|$                         & 57                \\
    $\DualAgentModel$ Observation dim $|\hat{o}_i|$                     & 114               \\
    $\SingleAgentModel$ Action dim $|\DeltaQ_i|$                        & 6                 \\
    $\DualAgentModel$ Action dim $|\DeltaQ_i|$                          & 6                 \\
    Policy learning rate                                                & 1e-4              \\
    Learning rate weight decay                                          & 1e-6              \\
    Polyak update coefficient (EMA decay)                               & 0.001             \\
    Optimizer                                                           & AdamW             \\ \midrule
    \multicolumn{2}{@{}l}{\textit{Planning \& Cost Function}}                               \\
    Planning batch size $B$ (candidate samples per arm)                 & 10                \\
    Planning timeout                                                    & 60s               \\
    Workspace radius (defines task difficulty)                          & 0.85              \\
    Plan segment collision penalty $P_{\text{coll}}$                    & 10                \\
    \bottomrule
  \end{tabular}
\end{table*}

\subsection{Effect of value-based diffusion models}

\begin{wrapfigure}{r}{0.5\textwidth}
\begin{minipage}{0.48\textwidth}
\centering
\begin{tabular}{@{}ccc@{}}
  \toprule
  \textbf{Method} & \textbf{Success} $(\uparrow)$ & \textbf{Steps} $(\downarrow)$ \\
  \midrule
  DG-MAP ($\SingleAgentModel, \DualAgentModel$)     & 0.890 & 5390 \\
  DG-MAP ($\SingleAgentModelQL, \DualAgentModelQL$) & \textbf{0.908} & \textbf{5254} \\
  \bottomrule
\end{tabular}
\captionof{table}{Success rate (\%, higher is better) along with average number of steps (lower is better) taken by the DG-MAP variants to complete the full cycle of multi-arm pick-and-place task.}
\label{tab:diffql_pnp}
\end{minipage}
\end{wrapfigure}

To address the challenge of out-of-distribution performance in diffusion models, the DiffusionQL \cite{wang2023diffusionpoliciesexpressivepolicy} method was recently introduced. Motivated by this, we trained two new models, $\SingleAgentModelQL$ and $\DualAgentModelQL$, to test the effectiveness of DiffusionQL within our planning framework. First, we generated offline RL datasets compatible with DiffusionQL training. This was done by taking the data collected for our original single-arm and dual-arm models and ``retracing'' it through the environment simulator, applying the reward function described in \cite{ha2020multiarm}. Using these offline RL datasets, we then adapted the DiffusionQL critic for use in a receding horizon context. Specifically, we modified the critic to evaluate entire action sequences over the full prediction horizon. Finally, during the planning stage, we sampled 50 candidate action sequences per arm from the trained DiffusionQL actors. We subsequently employed the adapted critic to score these sequences and selected the top 10 sequences for each arm. These selected sequences were then used as input for the remainder of our established planning algorithm.

\begin{wrapfigure}{r}{0.55\textwidth}
\vspace{-10pt}
\centering
\begin{tabular}{@{}ccccc@{}}
\toprule
\multirow{2}{*}{\textbf{Arms}} & \multicolumn{4}{c}{\textbf{Average}}            \\ \cmidrule(l){2-5} 
 &
  \begin{tabular}[c]{@{}c@{}}Baseline\\ LD\end{tabular} &
  \begin{tabular}[c]{@{}c@{}}Baseline\\ ED\end{tabular} &
  \begin{tabular}[c]{@{}c@{}}DG-MAP\\ ($\SingleAgentModel, \DualAgentModel$)\end{tabular} &
  \begin{tabular}[c]{@{}c@{}}DG-MAP\\ ($\SingleAgentModelQL, \DualAgentModelQL$)\end{tabular} \\ \midrule
3 & 0.412 & 0.965 & \textbf{0.975} & 0.973 \\
4 & 0.039 & 0.961 & 0.968          & \textbf{0.970} \\
5 & 0.075 & 0.934 & 0.945          & \textbf{0.955} \\
6 & 0.055 & 0.906 & \textbf{0.928} & 0.923 \\
7 & 0.038 & 0.917 & \textbf{0.926} & 0.924 \\
8 & 0.022 & 0.907 & \textbf{0.924} & 0.919 \\
\bottomrule
\end{tabular}
\vspace{-5pt}
\makeatletter\def\@captype{table}\makeatother
\caption{\textbf{Average Success Rates}(\%, higher is better) for 3–8 arms across different methods}
\label{tab:diffql}
\end{wrapfigure}

Table~\ref{tab:diffql_pnp} presents results on the multi-arm pick-and-place task. Here, the DG-MAP variant utilizing DiffusionQL models achieves a slightly higher success rate (90.8\% vs. 89.0\%) and completes the task in marginally fewer steps on average (5254 vs. 5390) compared to the standard variant. This suggests that incorporating reward signals via Q-learning might offer a small advantage in optimizing for success and efficiency on this complex, multi-stage application task, potentially by better handling scenarios that deviate from the expert demonstrations.

However, examining the average success rates across the general goal-reaching benchmarks (Table~\ref{tab:diffql}), the performance difference between the two DG-MAP variants is minimal and inconsistent across different numbers of arms. For $N=4, 5$, the QL variant shows a slight edge, while for $N=3, 6, 7, 8$, the standard variant performs marginally better. The absolute difference in average success rate is typically less than 1\% between the two. This indicates that for the broad set of goal-reaching tasks, both the behavioral cloning and Diffusion Q-Learning approaches provide highly effective underlying generative models for trajectory proposals and repairs within the DG-MAP planning framework. The core benefit appears to stem from the planner's structure combined with the generative capabilities of diffusion models, rather than a strong preference for either the standard or QL training objective in the general case.
\end{document}